\documentclass[conference]{IEEEtran}
\pdfoutput=1

\IEEEoverridecommandlockouts
\IEEEpubid{\makebox[\columnwidth]{978-1-6654-8045-1/22/\$31.00~\copyright2022 IEEE \hfill}
\hspace{\columnsep}\makebox[\columnwidth]{ }}

\usepackage[utf8]{inputenc}
\usepackage[T1]{fontenc}
\usepackage{verbments}
\usepackage[spanish, english]{babel} 

\usepackage{amsmath,amssymb,amsfonts}
\usepackage{algorithmic}
\usepackage{textcomp}
\usepackage{xcolor}

\usepackage{times}
\usepackage{epsfig}
\usepackage{graphicx}
\usepackage{amsmath}
\usepackage{amssymb}
\usepackage{enumitem}

\usepackage{balance}
\usepackage[numbers, sort&compress]{natbib}

\usepackage{longtable}
\usepackage{adjustbox}
\usepackage{multirow}
\usepackage{tablefootnote}
\usepackage{pifont}
\usepackage{booktabs}
\newcommand{\cmark}{\ding{51}}%
\newcommand{\xmark}{\ding{55}}%

\usepackage[pagebackref=false,breaklinks=true,colorlinks,bookmarks=false]{hyperref}

\makeatletter
\newcommand{\linebreakand}{%
  \end{@IEEEauthorhalign}
  \hfill\mbox{}\par
  \mbox{}\hfill\begin{@IEEEauthorhalign}
}
\makeatother

\def\BibTeX{{\rm B\kern-.05em{\sc i\kern-.025em b}\kern-.08em
    T\kern-.1667em\lower.7ex\hbox{E}\kern-.125emX}}
\begin{document}

\title{Towards Automatic Cetacean Photo-Identification: A Framework for Fine-Grain, Few-Shot Learning in Marine Ecology}

\author{
    \IEEEauthorblockN{
                        Cameron Trotter\IEEEauthorrefmark{1},
                        Nick Wright\IEEEauthorrefmark{1},
                        A. Stephen McGough\IEEEauthorrefmark{2},
                        Matt Sharpe\IEEEauthorrefmark{3},\\
                        Barbara Cheney\IEEEauthorrefmark{4},
                        M\`{o}nica Arso Civil\IEEEauthorrefmark{5},
                        Reny Tyson Moore\IEEEauthorrefmark{6},
                        Jason Allen\IEEEauthorrefmark{6}, and
                        Per Berggren\IEEEauthorrefmark{3}}\\
    \IEEEauthorblockA{
                        \IEEEauthorrefmark{1}School of Engineering, \IEEEauthorrefmark{2}School of Computing, \IEEEauthorrefmark{3}School of                          Natural \& Environmental Sciences\\
                                            Newcastle University, UK\\
                                            \{c.trotter2, nick.wright, stephen.mcgough, m.j.sharpe, per.berggren\}@ncl.ac.uk}\\

    \IEEEauthorblockA{
        \begin{tabular}{cc}
            \begin{tabular}{@{}c@{}}
                \IEEEauthorrefmark{4}School of Biological Sciences\\
                                            University of Aberdeen, UK\\
                                            b.cheney@abdn.ac.uk\\
            \end{tabular} & \begin{tabular}{@{}c@{}}
                                        \IEEEauthorrefmark{5}School of Biology\\
                                            University of St Andrews, UK\\
                                            mac64@st-andrews.ac.uk
            \end{tabular}
        \end{tabular}
    }\\
    \IEEEauthorblockA{
                    \IEEEauthorrefmark{6}Chicago Zoological Society's Sarasota Dolphin Research Program\\
                                        Sarasota, FL, USA\\
                                        renytysonmoore@gmail.com, allenjb@mote.org}
}

\maketitle
\vspace*{-10mm}
\begin{abstract}
   Photo-identification (photo-id) is one of the main non-invasive capture-recapture methods utilised by marine researchers for monitoring cetacean (dolphin, whale, and porpoise) populations. This method has historically been performed manually resulting in high workload and cost due to the vast number of images collected. Recently automated aids have been developed to help speed-up photo-id, although they are often disjoint in their processing and do not utilise all available identifying information. Work presented in this paper aims to create a fully automatic photo-id aid capable of providing most likely matches based on all available information without the need for data pre-processing such as cropping. This is achieved through a pipeline of computer vision models and post-processing techniques aimed at detecting cetaceans in unedited field imagery before passing them downstream for individual level catalogue matching. The system is capable of handling previously uncatalogued individuals and flagging these for investigation thanks to catalogue similarity comparison. We evaluate the system against multiple real-life photo-id catalogues, achieving mAP@IOU[0.5] = 0.91, 0.96 for the task of dorsal fin detection on catalogues from Tanzania and the UK respectively and 83.1, 97.5\% top-10 accuracy for the task of individual classification on catalogues from the UK and USA.
\end{abstract}


\section{Introduction}\label{sec:intro}

In recent years there has been a concerted effort to apply computer vision techniques to challenging big data problems which can have a positive societal impact. A highly important area where computer vision can help is ecology \cite{weinstein_computer_2018}. One of the main goals of ecological research is to monitor animal populations in their distribution area, undertaking abundance estimates to inform policy change. This is most commonly performed using capture-recapture surveys where researchers identify the presence of individuals and estimate abundance of animals in an area to produce population estimates \cite{sharpe_indian_2019, arso_civil_changing_2019, cheney_long-term_2014}. These surveys can be classified as invasive where animals are physically trapped, tagged, and released, or non-invasive where monitoring is performed passively such as via the collection of images -- referred to as photo-id. 

Photo-id is one of the main non-invasive capture-recapture methods utilised by cetacean researchers \cite{hammond_individual_1990}. Surveys are usually undertaken at sea, although monitoring from coastlines or aircraft may also be utilised \cite{payne_long_1986, forney_seasonal_1998}. The methodology is employed for the monitoring of multiple cetacean species, with a range of studies demonstrating its efficacy \cite{feyrer_origin_2021, bigg_assessment_1982}.

\begin{figure}
	\begin{center}
		\includegraphics[width=\linewidth]{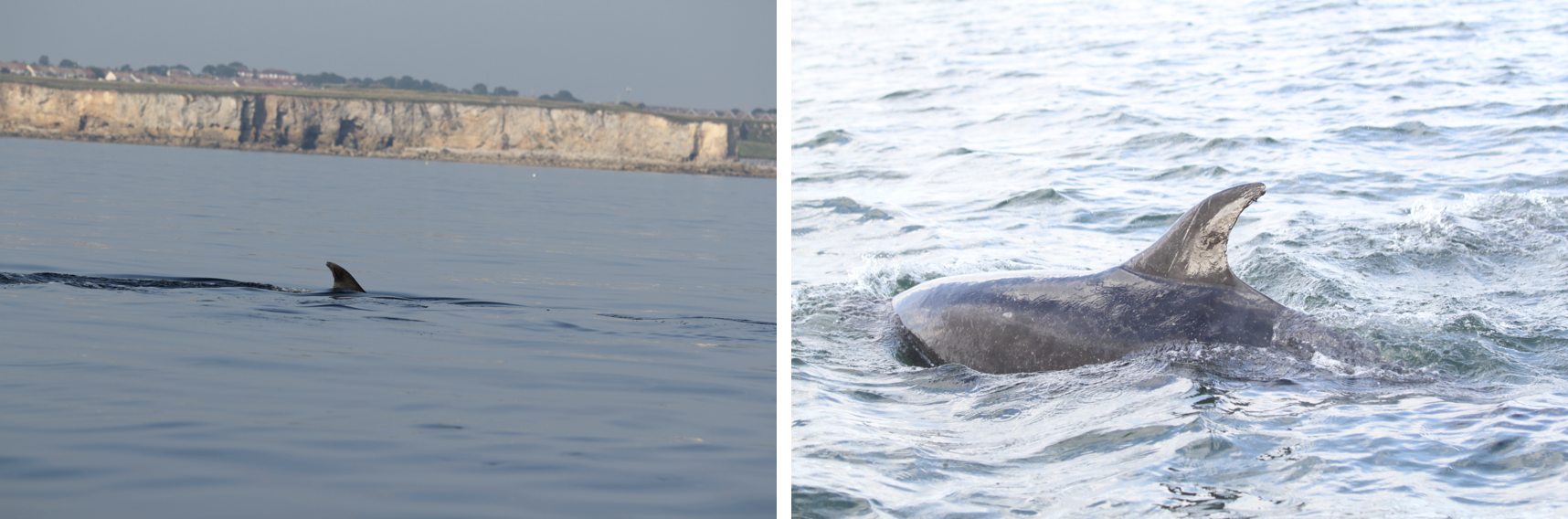}
	\end{center}
	\vspace*{-4mm}
	\caption{Two images of the same individual taken from different angles of approach, directions of travel, surfacing elevation, and distances from the camera. This changes the make-up of the dorsal fin but retains identifying information. Images from \cite{trotter_ndd20_2020}.}
	\label{fig:angle-size-example}
	\vspace*{-5mm}
\end{figure}

All non-invasive capture-recapture methodologies rely on the target species having some form of individually identifiable markings. Depending on the species, different parts of the body are the primary identifying feature; for dolphins this is usually the dorsal fin as this body part is most likely to be visible above the waterline. During photo-id surveys, researchers often focus on long lasting stable markers such as dorsal fin shape, notches, scarring, and pigmentation. These markings can be difficult to capture in detail due to the free roaming nature of the animals causing high variances in angles of approach, direction of travel, distance from camera, and surfacing elevation, as seen in Figure \ref{fig:angle-size-example}. This is exacerbated when dealing with cetacean species which travel in pods, making it difficult to distinguish the individuals present.

Marine photo-id can be extremely labour and cost intensive compared to on-land surveys, which rely on the use of camera traps placed in stationary locations to capture images when they detect movement. This setup is not possible at sea due to a lack of stationary objects to attach devices to and rapid movement in the observed scene due to waves causing the camera to trigger -- producing a high false positive rate.

 Upon survey completion, photo-id data must be analysed and individuals identified to produce a catalogue. Images collected during surveys are large in size and contain significant amounts of background noise. Historically curation of this data has been a manual process that often takes longer than the entire data collection period \cite{tyson_moore_rise_2022}, further increasing labour and costs. As such, any techniques to speed up the curation process would be welcomed by both researchers and their funding bodies. As photo-id surveys are not guaranteed to capture all individuals in a given geographic area, naive approaches such as training a simple image classifier on existing catalogue examples would not suffice as they are incapable of flagging previously uncatalogued individuals.

This work details a framework for fully automatic catalogue matching based on unprocessed photo-id imagery. This is achieved by a pipeline of trained computer vision models and robust post-processing techniques capable of automatic fin detection and most likely catalogue matching based on latent space similarity. Images are passed through a Mask R-CNN \cite{he_mask_2017} dorsal fin detector, removing the need for manual data pre-processing. Detections are post-processed ready for fine-grain, few-shot catalogue matching via a Siamese Neural Network (SNN) trained using triplet loss \cite{schroff_facenet_2015} and online semi-hard triplet mining to create a latent space based on the provided catalogue. Matches are obtained using the Euclidean distances between an input and generated class prototypes, allowing for the flagging of potentially uncatalogued individuals. This reduction in data processing affords cetacean researchers more time to work on application of their data, for example to inform mitigation and policy change, rather than curation. 


\section{Related Work}\label{sec:relatedWork}

Due to the time and labour requirements of manual photo-id, multiple aids have been developed. Descriptions of these related works is provided, with an overview in Table \ref{tab:Photo-IDAidsComparison}.

\begin{table*}
	\caption[A comparison of available photo-id aids.]{A comparison of available photo-id aids.}\label{tab:Photo-IDAidsComparison}
	\begin{adjustbox}{width=\linewidth, center}
		\begin{tabular}{ccccccc}
            \hline
            System                                                                                                  & \begin{tabular}[c]{@{}c@{}}Requires\\Data Pre-processing\end{tabular} & \begin{tabular}[c]{@{}c@{}}Dorsal Fin\\Detection\end{tabular}             & \begin{tabular}[c]{@{}c@{}}Full\\ Background Removal\end{tabular} & \begin{tabular}[c]{@{}c@{}}Individual\\Photo-ID\end{tabular}                         & \begin{tabular}[c]{@{}c@{}}Can Flag Individuals\\Not Currently In Catalogue\end{tabular} & \begin{tabular}[c]{@{}c@{}}Uses All Information\\Found on Dorsal\end{tabular} \\ \hline
            FinScan \cite{hillman_finscan_2002}                                                            & \cmark                                                           & \xmark & \xmark                                             & \cmark            & \cmark                                                & \textthreequartersemdash                                  \\
            FinBase \cite{fisheries_finbase_2018}                                                          & \cmark                                                           & \xmark & \xmark                                             & \cmark            & \cmark                                                & \textthreequartersemdash                                  \\
            DARWIN \cite{hale_unsupervised_2012}                                                           & \cmark                                                           & \xmark & \xmark                                             & \cmark            & \cmark                                                & \textthreequartersemdash                                  \\
            catRlog \cite{keen_catrlog_2021}                                                               & \cmark                                                           & \xmark & \xmark                                             & \cmark            & \cmark                                                & \textthreequartersemdash                                  \\
            CurvRank \cite{weideman_integral_2017}                                                        & \cmark$^*$                                                           & \xmark$^*$ & \xmark                                             & \cmark            & ?                                                                    & \xmark                                  \\
            Karnowski \textit{et al.} \cite{karnowski_dolphin_2015}                       & \xmark                                                           & \cmark & \cmark                                             & \xmark            & \textthreequartersemdash                              & \textthreequartersemdash                                  \\
            Photo-ID Ninja\textsuperscript{\ref{footnote:photo-idNinja}}                    & \xmark                                                           & \cmark & \xmark                                             & \xmark            & \textthreequartersemdash                              & \textthreequartersemdash                                  \\
            Qui\~{n}onez \textit{et al.} \cite{quinonez_using_2019} & \xmark                                                           & \cmark & \xmark                                             & \xmark            & \textthreequartersemdash                              & \textthreequartersemdash                                  \\
            Morteo \textit{et al.} \cite{morteo_phenotypic_2017}                          & \cmark                                                           & \xmark & \xmark                                             & \cmark            & \xmark                                                & \xmark                                                    \\
            Bouma \textit{et al.} \cite{bouma_individual_2018}                            & \cmark                                                 & \cmark$^\dagger$ & \xmark                                             & \cmark            & \cmark                                                & \cmark                                                    \\
            Lee \textit{et al} \cite{lee_backbone_2020}                                   & \xmark                                                           & \cmark & \cmark                                             & \cmark & ?                                                                    & \textthreequartersemdash                                  \\
            finFindR \cite{thompson_finFindR_2022}                                                         & \xmark                                                           & \cmark & \xmark                                             & \cmark            & \cmark                                                & \xmark                                                    \\
            DolFin \cite{maglietta_dolfin_2018}                                             & \xmark                                                           & \cmark & \cmark                                             & \cmark$^\mathsection$            & \cmark                                                & \cmark                                                    \\ \hline
            Ours                                                                                                              & \xmark                                                           & \cmark & \cmark                                             & \cmark            & \cmark                                                & \cmark                                                    \\ \hline
        \end{tabular}
	\end{adjustbox}
	{\raggedright\footnotesize {* CurvRank is included in Flukebook (\href{http://flukebook.org}{flukebook.org}), a website for cross-catalogue matching. In this scenario, Flukebook performs automatic data pre-processing and fin detection before passing to CurvRank, but the algorithm itself does not facilitate this. \\ ? It is unclear whether the system is capable of flagging previously uncatalogued individuals to the researcher based on the literature available. \\ $\dagger$ Utilises Photo-ID Ninja for detection. \\ $\mathsection$ Utilises SURF for photo-id, thus unsuitable for cetacean species without well-defined markings.} \par}
	\vspace*{-4mm}
\end{table*}

\begin{description}[wide, itemindent=\labelsep]

\item[Catalogue Management Systems] are widely used to aid manual photo-id, especially in geographic locations with large resident populations. FinScan \cite{hillman_finscan_2002} allows users to upload pre-processed fin images which they then trace around. This trace is checked against a database to determine most likely matches. Likewise, DARWIN \cite{hale_unsupervised_2012} also provides automated photo-id based on traces.

FinBase \cite{fisheries_finbase_2018} instead manages catalogues based on user defined attributes which can then be used for matching. Fins are matched based on querying a database for entries with matching attributes. Additionally, CatRlog \cite{keen_catrlog_2021} provides likely matches based on manually entered marking information.

\item[Non-Deep Learning Approaches] to aid cetacean photo-id are available. Karnowski \textit{et al.} \cite{karnowski_dolphin_2015} used PCA to subtract background from underwater images of common bottlenose dolphins (\textit{Tursiops truncatus}). Further, Weideman \textit{et al.} \cite{weideman_integral_2017} proposed CurvRank, an algorithm which automatically identifies a fin's trailing edge for likely matching.

\item[Deep Learning Approaches] have also been explored in recent years, inching work towards a fully automatic photo-id system. Data pre-processing such as cropping is one of the main labour costs in catalogue creation. Photo-ID Ninja\footnote{\label{footnote:photo-idNinja}Photo-ID Ninja: \href{http://photoid.ninja}{photoid.ninja}} aims to speed up this processing by automatically cropping images to the dorsal fin.

Qui\~{n}onez \textit{et al.} \cite{quinonez_using_2019} proposed a detection system capable of distinguishing between four distinct classes: \texttt{dolphin}, \texttt{dolphin\_pod}, \texttt{open\_sea}, and \texttt{seabirds}. Morteo \textit{et al.} \cite{morteo_phenotypic_2017} performed semi-automatic matching by projecting lines from the base of the fin's leading edge.

Bouma \textit{et al.} \cite{bouma_individual_2018} provided a system focusing on metric learning to photo-id individual New Zealand common dolphins (\textit{Delphinus spp}), utilising Photo-ID Ninja to crop fins prior to matching. Lee \textit{et al}. \cite{lee_backbone_2020} proposed an architecture for cetacean identification via normalised segmentation and significant feature extraction.

DolFin \cite{maglietta_dolfin_2018} uses a SURF-based \cite{bay_speeded-up_2008} approach to identify Risso's dolphins (\textit{Grampus griseus}), a species prone to long term scarring which makes them ideal for matching via feature extractors. This approach fails with species where identifying markings are more subtle, such as in Figure \ref{fig:SURF-Failure} where SURF has failed to extract the top left notch from the dorsal fin of an Indo-Pacific bottlenose dolphin (\textit{T. aduncus}). 

\begin{figure}
    \vspace*{-6mm}
	\begin{center}
		\includegraphics[width=0.3\linewidth]{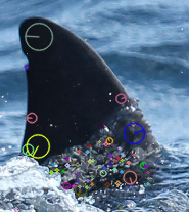}
	\end{center}
	\vspace*{-4mm}
	\caption{An example showing the result of SURF \cite{bay_speeded-up_2008} feature extraction on the dorsal fin of an Indo-Pacific bottlenose dolphin.}
	\label{fig:SURF-Failure}
	\vspace*{-6mm}
\end{figure}

FinFindR \cite{thompson_finFindR_2022} allows for the identification of common bottlenose dolphins based on the trailing edge of the animal's dorsal fin, utilising this to cluster individuals.
\end{description}

\subsection{Comparison Against Related Work}\label{sec:relatedWork,sub:Comparison}

Table \ref{tab:Photo-IDAidsComparison} provides a summary of related works. The vast majority are standalone algorithms, requiring researchers to set up their own data pipelines. Only DolFin and finFindR propose fully automated, self-contained photo-id aids which require no pre-processing and are capable of handling previously uncatalogued individuals.

As noted previously, DolFin may struggle to handle species other than Risso's dolphin as SURF fails to extract subtle features such as notches. Further, finFindR fails to make use of all information on the dorsal, utilising only the trailing edge for matching, and does not fully remove background noise which may influence the matching process.

Work proposed in this paper aims to overcome these limitations. The outlined methodology performs full background removal, negating the effect of noise on matching, and is capable of operating on a range of species rather than just those prone to specific markings by extracting all available information. Images can also be operated over without the need for manual data pre-processing.


\section{Methodology}\label{sec:methodology}

\begin{figure*}
	\begin{center}
		\includegraphics[width=\linewidth]{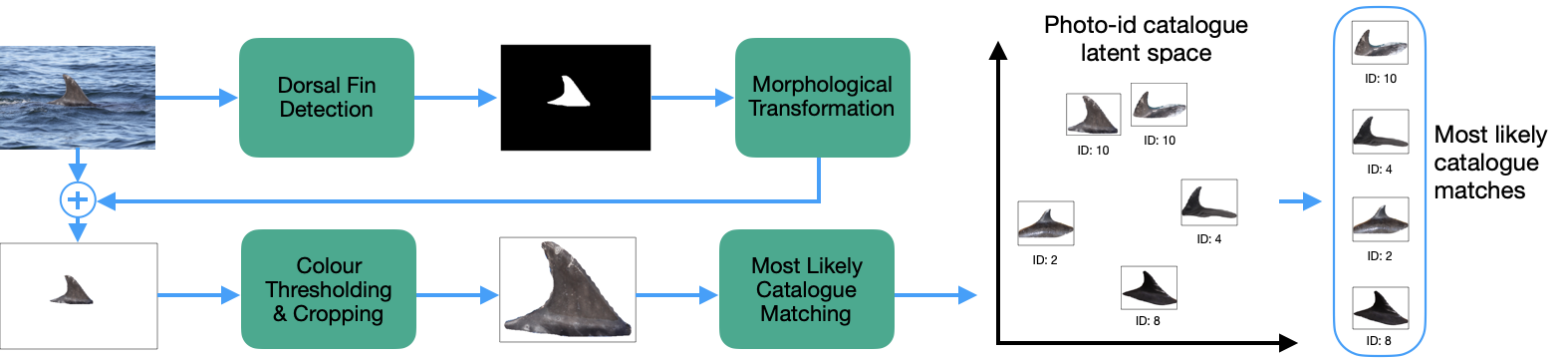}
	\end{center}
	\caption{A high level overview of data flow through the proposed system.}
	\label{fig:pipeline}
	\vspace*{-5mm}
\end{figure*}

The proposed framework allows for the detection and individual identification of various cetacean species. This is achieved through a pipeline of models (see Figure \ref{fig:pipeline}) to sequentially process input images.

\begin{description}[wide, itemindent=\labelsep]

\item[Dorsal Fin Detection] is achieved using a Mask R-CNN model to locate regions of interest (RoIs) in images, defined as areas where the animal's dorsal fin is visible above the waterline. This model is trained on large scale images (3456x5184px) from DSLR cameras, typical of those utilised during photo-id surveys. By detecting RoIs in input images automatically, the requirement for data pre-processing is removed. Model outputs are in the form of masks which precisely differentiate between a coarse-grain dorsal fin or background. If an image contains a pod ($>1$ dorsal fin), each detected RoI is processed sequentially downstream.

\item[Morphological Transformations] are performed based on \textit{a priori} knowledge of cetaceans. It can be reasoned that holes in a mask are likely unintentional and a product of surrounding noise. In this instance, the model may have failed to capture all available information. Any holes present in masks are filled using dilation and erosion morphological transformations to ensure no identifiable information is lost. Note that any holes present in the dorsal fin from natural or anthropogenic activity such as from sting ray barbs are not transformed to retain identifying information. A \textit{bitwise-and} operation is then applied between the clean mask and the input image, segmenting the RoI to reduce the amount of noise passed downstream. 


\item[Colour Thresholding] is executed over masks with multiple disjoint components. This may occur if, for example, an area of splash has been erroneously included as part of a detection. As a single cetacean cannot be made up of multiple disjoint components, it is known that some of these are erroneous and should be removed.

As the outer layer of cetaceans' skin is often a consistent grey colouring, minus any prominent identifiable markings, this can be used to filter mask components. By comparing the colour composition of each component against a calculated threshold, it is possible to discard those which have been erroneously detected. As these components are often areas of water, they will likely be much lighter in composition than the cetacean component. An example of noise removal via colour thresholding is shown in Figure \ref{fig:thresholding-eg}. If multiple mask components pass noise removal and colour thresholding, each component is treated as a distinct mask downstream. Masks with only a single component are not colour thresholded to ensure no detections are ignored due to post-processing, preventing the discarding of an RoI which contains no disjoint components but is above the threshold, such as in the event of extreme over-exposure.

\begin{figure}[h]
	\begin{center}
		\includegraphics[width=\linewidth]{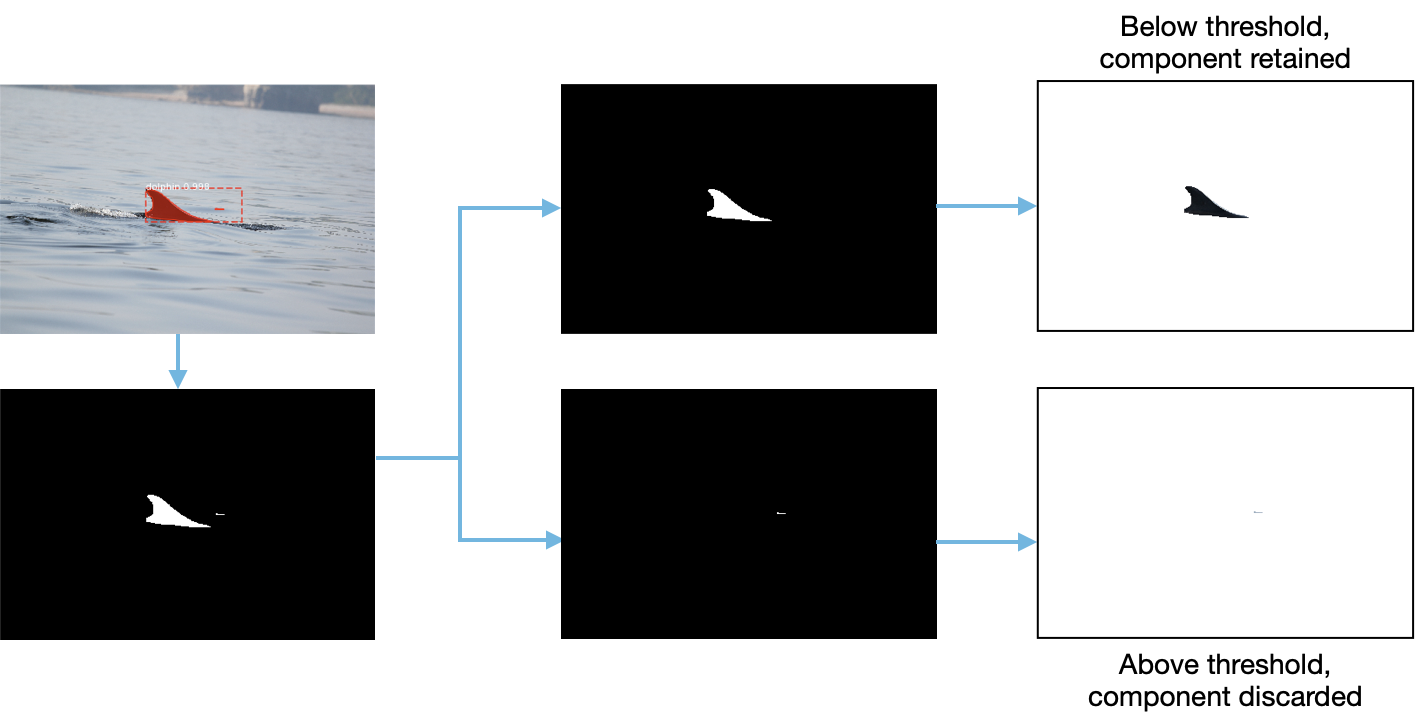}
	\end{center}
	\vspace*{-4mm}
    \caption{Workflow detailing colour thresholding to remove disjoint splash. The mask is split into individual components. The background subtracted images are colour thresholded, and erroneous splash is discarded.}
    \label{fig:thresholding-eg}
	\vspace*{-7mm}
\end{figure}

\item[Cropping] is then undertaken, reducing the input image down to the processed RoI. This vastly reduces the image file size and computational expense of downstream operations, as well as centres the RoI in the output image.


\item[Most Likely Catalogue Matching] is applied on a per-catalogue basis using a trained SNN with a triplet loss function \cite{schroff_facenet_2015}. SNNs have demonstrated usefulness for few-shot, fine-grain problems in research domains such as species identification \cite{vetrova_hidden_2018, araujo_two-view_2022}. Our work compliments this through extension to individual-level identification.

At train time, the photo-id catalogue is processed as per the methodology above. Each post-processed RoI for an individual is used as a class example for SNN training. A \texttt{noise} class is also included, containing examples of retained erroneous detections, to afford the model the ability to rule these out during inference without human intervention.

The trained SNN is capable of generating low-dimensional embeddings from fin images in such a way that those of the same individual generate embeddings which are close together in the latent space, creating individual class clusters which allow for most likely catalogue matching. During inference, most likely matching is performed via Euclidean distance measurement between the input and prototypes representing the median example of each class.
 
By clustering individuals together in the latent space based on similarity and comparing new images to the class prototypes, the system can flag potentially previously uncatalogued individuals. A new individual who enters the catalogue's survey area (e.g. through migration or birth) is placed in a distinct latent space location, resulting in large distances between it and the class prototypes. If a new individual is flagged it can then be verified by a human and, if correct, added to the catalogue. When a new entry is appended, a class prototype for the newly introduced individual can be generated, allowing the system to perform catalogue matching without the need for model re-training. 
\end{description}


\section{Experimentation}\label{sec:Experimentation}

To evaluate the proposed methodology, experimentation was performed using multiple real-life photo-id catalogues collected by various institutions from a range of geographic locations, times, and encompassing multiple cetacean species.

\subsection{Dorsal Fin Detection}\label{sec:Experimentation,sub:finDetection}

\begin{description}[wide, itemindent=\labelsep]

\item[Precision] 

The dorsal fin detector's ability to precisely detect RoIs was evaluated using mean average precision at differing intersection over union thresholds (mAP@IOU). This was performed using 1021 photo-id survey images provided by Newcastle University's Marine MEGAfauna Lab of Indo-Pacific bottlenose dolphins resident in the coastal waters of Zanzibar, Tanzania in 2015 \cite{sharpe_indian_2019}. RoIs were labelled with a coarse-grain \texttt{dolphin} class, resulting in 616 total examples split into an 80-20 train-test split.

A ResNet50 architecture \cite{he_deep_2015} was utilised as a model backbone, with a minimum confidence threshold of 0.9. Hyperparameter optimisation was performed using a grid search, with the following possibilities examined: (1) \textit{Weight Decay}: 0.01, 0.001, 0.0001, or 0.0001. (2) \textit{RPN Anchor Scale}: (8, 16, 32, 64, 128), (16, 32, 64, 128, 256), or (32, 64, 128, 256, 512). (3) \textit{Optimiser}: Adam \cite{kingma_adam:_2014}, or SGD with Warm Restarts \cite{loshchilov_sgdr:_2016}. (4) \textit{Pre-trained on MSCOCO} \cite{lin_microsoft_2014}: yes or no. (5) \textit{Data Augmentation Strategy}: \texttt{aug1}, \texttt{aug2}, or None. 

The first strategy, \texttt{aug1}, selected up to three of the following: (1) \textit{Horizontal Flip}: (p = 0.5). (2) \textit{Vertical Flip}: (p = 0.5). (3) \textit{Rotation}: 90, 180, or 270 degrees (p = 0.33). (4) \textit{Scaling}: 80\% to 120\% on both axes independently. (5) \textit{Brightness}: multiply all pixels in the image with a random value between 0.8 and 1.5. (6) \textit{Gaussian Blur}: using a kernel with radius randomly assigned between 0 and 5. 

The second strategy, \texttt{aug2}, was more complex, performing the following perturbations in a sequentially random order on 67\% of the images: (1) \textit{Horizontal Flip}: (p = 0.5). (2) \textit{Cropping}: each side of the image randomly between 0\% and 10\% of the total side length. (3) \textit{Gaussian Blur}: using a kernel with radius randomly assigned between 0 and 2.5 (p = 0.5). (4) \textit{Contrast}: increase or decrease by a random factor between 0.75 and 1.5. (5) \textit{Additive Gaussian Noise}: sample the noise per channel, adding noise to the colour of the pixels. (6) \textit{Brightness}: multiply all pixels with a random value between 0.8 and 1.2. (7) \textit{Scaling}:  80\% to 120\% on both axes independently. (8) \textit{Rotation}: randomly between -180 and 180 degrees. The use of these augmentation strategies allowed for evaluation of whether a simple or more complex strategy would be more appropriate for this use case.

The search determined that training a model optimised using SGD with Warm Restarts, alongside an initial learning rate of 0.001 with 0.01 weight decay, RPN anchor scales of (16, 32, 64, 128, 256), pretrained on MSCOCO and using the \textit{aug1} strategy produced the highest mAP@IOU scores. 

Experimental results for the trained model on the Zanzibar data can be seen in Table \ref{tab:maskrcnn_results} (Top). These images contain a wide variety of background noise. Furthermore, some dorsal fins in the images look similar in shape and structure to background objects -- especially in choppy waters captured from a distance. The animal’s bodies are also similarly coloured to their surroundings. These adaptations allow the animals to be better camouflaged in their environment, but can cause issues for detection systems. The model is capable of precisely detecting multiple fins when the animals are captured travelling in a pod, as seen in Figure \ref{fig:pod-detection-and-boat} (Left) where the pod has been separated into its constituent parts ready for identification downstream.

\begin{figure}
	\begin{center}
		\includegraphics[width=\linewidth]{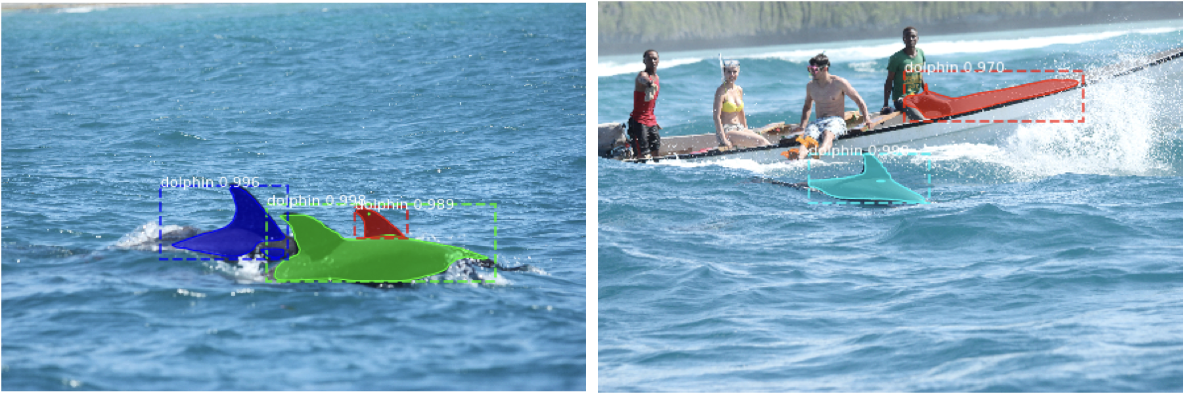}
	\end{center}
	\vspace*{-4mm}
    \caption{Example of dorsal fin detections. Left: An image showing cetaceans travelling in a pod. Each detection is post-processed and identified in isolation. Right: An image showing the effect of eco-tourism on the false positive rate of the detector.}
    \label{fig:pod-detection-and-boat}
\end{figure}

\item[Generalisability]

The detector is also required to produce detections with high mAP scores when operating on data from different geographic regions, time intervals, and species, as this would negate the need for detector re-training when working in a new survey area. To evaluate this, the model was used to generate mask predictions for the above water image set found in the Northumberland Dolphin Dataset (NDD) \cite{trotter_ndd20_2020}. This open-source dataset contains images of both common bottlenose dolphins and white-beaked dolphins (\textit{Lagenorhynchus albirostris}) collected during a 2019 photo-id survey off the coast of Northumberland, UK. 

\begin{table}
    \vspace*{-6mm}
    \caption{Instance segmentation results of the Mask R-CNN detector, trained using the Zanzibar training data.}
    \begin{adjustbox}{width=\linewidth, center}
        \begin{tabular}{ccccccccccc}
            \hline
            \multirow{2}{*}{\textbf{Dataset}} & \multicolumn{10}{c}{\textbf{mAP@IOU[$x$]}}                                                                                                                  \\
                                              & \textbf{0.50} & \textbf{0.55} & \textbf{0.60} & \textbf{0.65} & \textbf{0.70} & \textbf{0.75} & \textbf{0.80} & \textbf{0.85} & \textbf{0.90} & \textbf{0.95} \\ \hline
            \textbf{Zanzibar}                 & 0.91          & 0.91          & 0.89          & 0.86          & 0.85          & 0.79          & 0.69          & 0.50          & 0.15          & 0.00           \\
            \textbf{NDD}                      & 0.96          & 0.95          & 0.93          & 0.91          & 0.88          & 0.83          & 0.71          & 0.51          & 0.16          & 0.00          \\ \hline
        \end{tabular}
    \end{adjustbox}
	\label{tab:maskrcnn_results}
\end{table}

To test this generalisability, the model trained using the Zanzibar data was evaluated on the whole above water set of NDD without re-training or fine-tuning. Results for this can be seen in Table \ref{tab:maskrcnn_results} (Bottom). Interestingly, the model achieves a higher mAP at the given thresholds on NDD than the Zanzibar dataset on which it was trained. This is hypothesised to be due to the lack of other objects in NDD in comparison to the Zanzibar dataset. For example, some images in the Zanzibar dataset contain vessels as well as humans as a result of high levels of eco-tourism operating in the survey area \cite{christiansen_effects_2010}. This is not the case for the data collection area of NDD, which may lead to a reduction in the false positive rate of the model when evaluated on this dataset. Figure \ref{fig:pod-detection-and-boat} (Right) shows the effect of eco-tourism on the false positive rate, where the model believes a section of the boat's hull and the leg of a human to be a \texttt{dolphin} RoI. Regardless, this evaluation presents evidence that the model is robust enough to deal with data from a different geographic area, time, and cetacean species without the need for re-training or fine-tuning.
\end{description}

\subsection{Colour Thresholding Mask Components}\label{sec:Experimentation,sub:colourThresholding}

\begin{figure}
	\begin{center}
		\includegraphics[width=\linewidth]{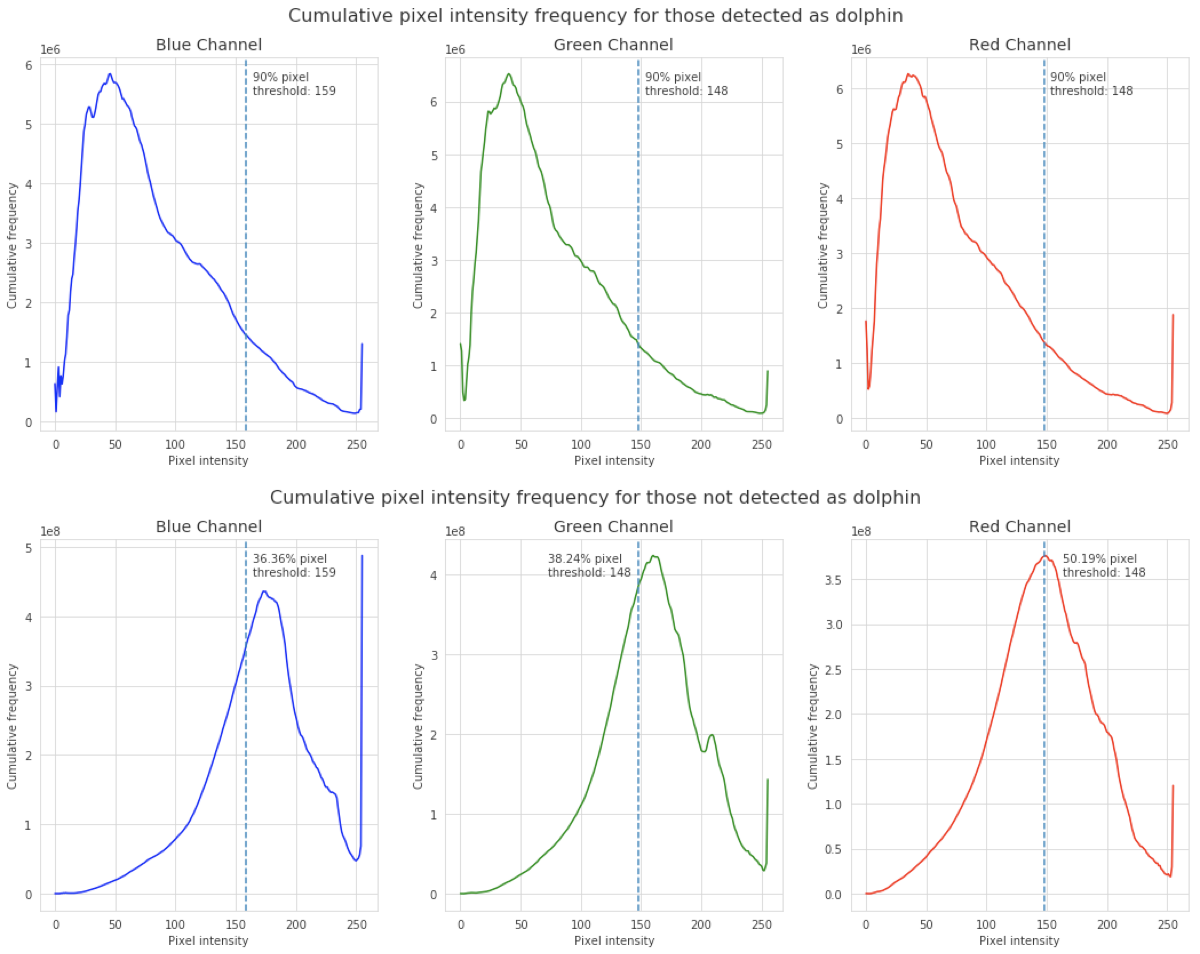}
	\end{center}
	\vspace*{-4mm}
	\caption{The global range of pixel intensities for each RGB colour channel of images in the Zanzibar dataset, split by pixel classification.}
	\label{fig:global-histogram}
	\vspace*{-5mm}
\end{figure}

Experimentation was undertaken to determine the optimal colour threshold value for mask post-processing. Histograms of the RGB colour channel pixel intensities for each object classification in the Zanzibar data were recorded, giving a total of six histograms per image. The histogram groups were then combined to give six global pixel intensity distributions, which can be seen in Figure \ref{fig:global-histogram}. Regardless of colour channel, there is a near inversion in the distribution of pixel intensities between those detected as \texttt{dolphin} and those not, strongly suggesting it is possible to determine if a component is erroneous based on its colour composition.

Globally, for all masks detected as \texttt{dolphin} 90\% of the RGB pixels are below intensities (148, 148, 159). As noise components in the mask are often areas of water or splash, these will be much lighter in composition than cetaceans, and thus can be removed from the mask with confidence by checking the percentage of pixels in the mask component below the threshold. 

It was found however that using a 90\% threshold when checking mask components at an individual image level was too restrictive, sometimes rejecting valid detections which may have been over-exposed due to lighting conditions. As such, whilst the colour threshold was kept the same, the percentage check was reduced to 50\% -- providing enough leeway for over-exposed but valid detections to be kept whilst still rejecting a large portion of erroneous ones. This was confirmed with further threshold testing on the NDD dataset.

\subsection{Most Likely Catalogue Matching}\label{sec:Experimentation,sub:MLM}

\begin{description}[wide, itemindent=\labelsep]

\item[Top-N Accuracy]

The SNN model was evaluated against its ability to clearly differentiate between classes in the latent space. To test this, an SNN was trained on the above water set of the NDD dataset after detection and post-processing.

Due to sparse amounts of example images for some individuals, additional photo-id data was provided from catalogues collected in waters around Eastern Scotland maintained by the University of Aberdeen and the University of St. Andrews Sea Mammal Research Unit \cite{arso_civil_changing_2019, cheney_long-term_2014}. Due to the large home range of cetaceans, a 23 individual overlap between this data and the NDD dataset was determined. As a result, 1827 additional images of the overlapping individuals were processed and included in the NDD dataset used to train the SNN. This consisted of 2626 images representing 44 classes including \texttt{noise}. Non-\texttt{noise} classes (median = 22) contain low inter-class but high intra-class differences between them (example images in Figure \ref{fig:snn-class-examples}). This provides a difficult few-shot, fine-grain dataset with which to evaluate the SNN's clustering ability. The dataset was divided using an 80-20 train-test split.

\begin{figure}
	\begin{center}
		\includegraphics[width=0.7\linewidth]{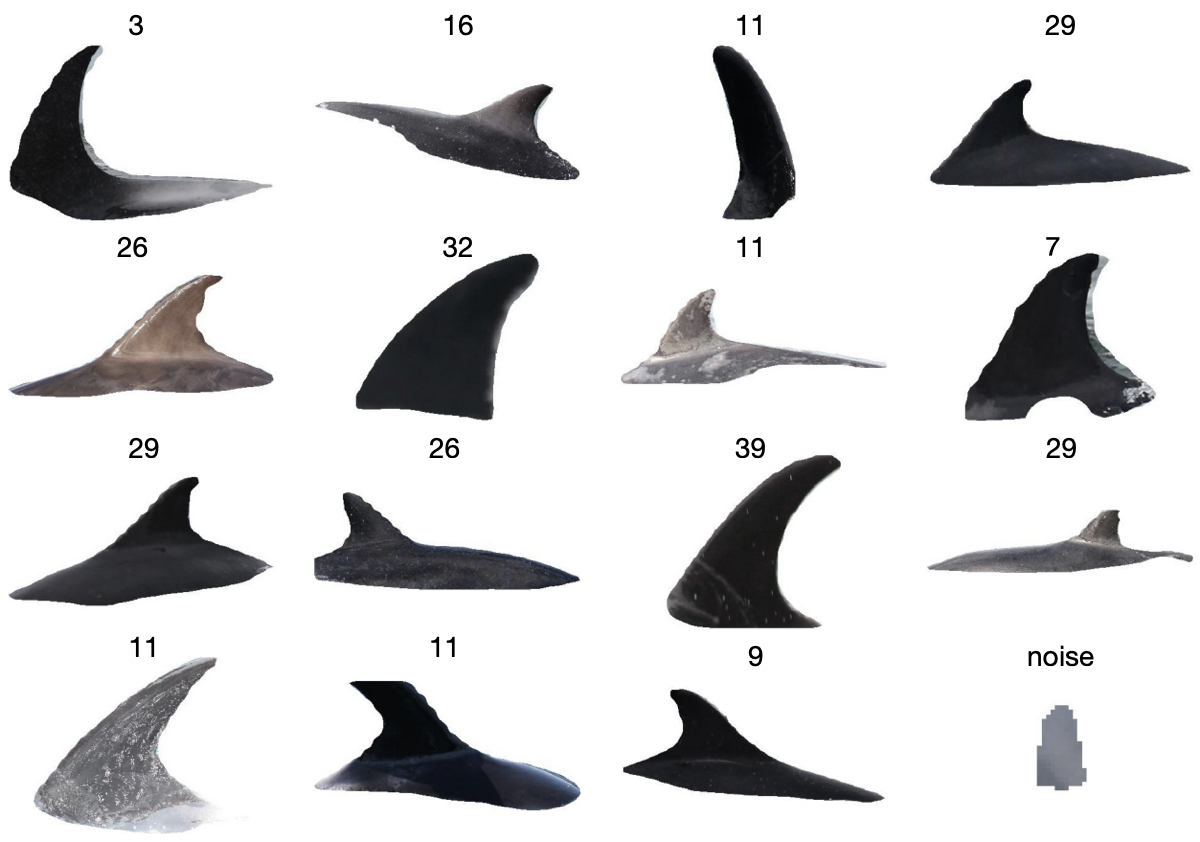}
	\end{center}
	\vspace*{-4mm}
	\caption{Example post-processed images from NDD used to train the SNN for the task of few-shot, fine-grain most likely photo-id catalogue matching. The class label is displayed above each image. Note the low inter-class and high intra-class differences between non-\texttt{noise} classes, such as between examples of class \texttt{11}.}
	\label{fig:snn-class-examples}
	\vspace*{-8mm}
\end{figure}

The SNN's backbone followed the structure outlined by Vetrova \textit{et al.} \cite{vetrova_hidden_2018}, with triplets selected via online semi-hard triplet mining. Hyperparameter optimisation of the SNN was performed using Optuna \cite{akiba_optuna_2019}. During the search, the following possibilities were examined: (1) \textit{Learning Rate} $\in \mathbb{R} \cap [1\times10^{-6}, 1\times10^{-3}]$, log uniform. (2) \textit{Dropout} $\in \mathbb{R} \cap [0.1, 0.7]$, log uniform. (3) \textit{Kernel Size} $\in \{5, 8\}$. (4) \textit{Triplet Loss Margin} $\in \mathbb{R} \cap [0.1, 1.0]$. (5) \textit{Weight Decay} $\in \mathbb{R} \cap [1\times10^{-6}, 1\times10^{-1}]$. (6) \textit{Optimiser} $\in \{SGD, Adam\}$. (7) \textit{Step Size} $\in \mathbb{Z} \cap [5, 10]$. (8) $\gamma \in \mathbb{R} \cap [0.1, 0.7]$, log uniform. (9) \textit{Embedding Size} $\in \mathbb{Z} \cap [16, 128]$.

Optimisation of the number of network blocks was also examined. Each block consisted of a Convolutional layer and a MaxPool layer (stride = 2), separated by a ReLU layer and a Dropout layer. During searching, the number of blocks was treated as a hyperparameter optimising for an \texttt{int} between 1 and 5 blocks. The initial Convolutional layer size was also tuned, searching for an optimal \texttt{int} value between 16 and 100. Subsequent layers were double the size of the previous.

The best performing hyperparameters were found to be an initial learning rate of $7.25\times10^{-6}$ with weight decay of 0.043 optimised using Adam, a kernel size of 6, triplet loss margin of 0.8, $\gamma$ of 0.012, dropout of 0.17, and an embedding size of 106 (giving a 106-dimensional latent space for most likely matching). The optimal number of network blocks was determined as 2, with an initial Convolutional layer size of 59. 

Various data augmentation strategies were examined. These were: (1) \textit{Colour Jitter}: randomly perturb the input images' brightness by a factor between 0.8 and 1.2, contrast by a factor between 0.8 and 1.2, saturation by a factor between 0.9 and 1.1, and hue by a factor between -0.1 and 0.1. (2) \textit{Perspective Shift}: randomly distort the input image's perspective by a factor of 0.5. (3) \textit{Both}: perform both \textit{Colour Jitter} and \textit{Perspective Shift}. (4) \textit{None}: no augmentations. 

Prototypes were generated based on the median of the embeddings generated for each class example. Test images were then processed by the SNN and their embedding plotted into the latent space, which was compared using Euclidean distance against the prototypes to generate a list of most likely matches utilised for top-$N$ accuracy evaluation. 

Results for the NDD dataset can be seen in Table \ref{tab:top_n_results} (Left). Best model performance was achieved without any data augmentation. These results confirm an SNN trained on automatically processed photo-id catalogue data is capable of accurately providing a list of most likely matches to cetacean researchers, even when trained only using a relatively small number of class examples.

\begin{table}
    \caption{Effect of different data augmentation strategies on SNN top-$N$ accuracy.}
    \begin{adjustbox}{width=\linewidth, center}
        \begin{tabular}{ccccccc}
            \cline{2-7}
            \textbf{}                                                                     & \multicolumn{3}{c}{\textbf{NDD}}                 & \multicolumn{3}{c}{\textbf{Naples}}               \\ \hline
            \textbf{\begin{tabular}[c]{@{}c@{}}Data Augmentation\\ Strategy\end{tabular}} & \textbf{Top-10} & \textbf{Top-5} & \textbf{Top-1} & \textbf{Top-10} & \textbf{Top-5} & \textbf{Top-1} \\ \hline
            \textit{Colour Jitter}                                                        & 76.83           & 61.18          & 38.82          & 96.25           & \textbf{91.25} & 70.00          \\
            \textit{Perspective Shift}                                                    & 73.58           & 51.22          & 23.17          & 96.25           & \textbf{91.25} & \textbf{73.75} \\
            \textit{Both}                                                                 & 78.46           & 61.18          & 40.04          & \textbf{97.50}  & 88.75          & 63.75          \\
            \textit{None}                                                                 & \textbf{83.13}  & \textbf{68.90} & \textbf{40.85} & 92.50           & 85.00          & 72.50          \\ \hline
        \end{tabular}
    \end{adjustbox}
    \vspace*{2mm}
	\label{tab:top_n_results}
	\vspace*{-7mm}
\end{table}

\item[Generalisability]

The generalisability of the SNN approach to most likely catalogue matching was evaluated using a photo-id catalogue subset provided by the Chicago Zoological Society's Sarasota Dolphin Research Program. The subset consisted of 250 images of 23 individual common bottlenose dolphins captured in the waters around Naples, FL, USA \cite{tyson_moore_final_2020}. Images were passed through the detector, post-processed, and the generated RoIs used to create a dataset capable of training an SNN for most likely catalogue matching. For consistency, the same architecture and hyperparameters were utilised as with the NDD dataset. 

Results for this dataset can be seen in Table \ref{tab:top_n_results} (Right). Unlike training on the NDD data where best results were achieved without any augmentation, here the results are more mixed. Whilst the best top-10 results are obtained using both \textit{Colour Jitter} and \textit{Perspective Shift} augmentations, the best top-5 results were obtained using only one strategy. Using \textit{Perspective Shift} only provided the best top-1 accuracy. Whilst this suggests that data augmentation strategy is catalogue dependent, the results confirm that accurate individual level most likely catalogue matching of cetaceans can be performed using SNNs trained on automatically pre-processed data.
\end{description}

\subsection{Effect of Background on Most Likely Catalogue Matching}\label{sec:Experimentation,sub:effectOfMasks}

Of the four works in Table \ref{tab:Photo-IDAidsComparison} which perform dorsal fin detection and downstream individual identification, only half remove all background beforehand. To examine the effect that background removal has on downstream identification, an SNN was trained using the NDD dataset processed into bounding box class examples. Due to the free roaming nature of the individuals, those in the NDD dataset were often photographed only during a single encounter leading to data with small intra-class but high inter-class background variation.

All variables, except the presence of background, were kept consistent with those used when training the best performing SNN on masked data, including model architecture, hyperparameters, and data augmentation strategy. Data was generated using the same Mask R-CNN detector as for previous experiments, modified to output bounding boxes rather than masks.

\begin{figure}
	\begin{center}
		\includegraphics[width=0.6\linewidth]{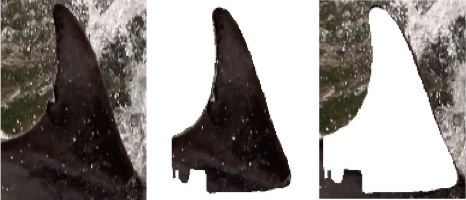}
	\end{center}
	\vspace*{-4mm}
	\caption{Example data used to examine the effect of retained background on most likely catalogue matching. Left: Bounding box detection containing both a dorsal fin and background. Centre: Corresponding dorsal fin mask. Right: Corresponding background mask.}
	\vspace*{-6mm}
	\label{fig:bboxvsmask}
\end{figure}

Analysis of the Euclidean distances between bounding boxed dorsal fins and corresponding fin and background masks show embedding generation is likely to be influenced more by features in the background than the fin. For example, the Euclidean distance between the bounding box data in Figure \ref{fig:bboxvsmask} (Left) and its corresponding dorsal fin mask (Centre) is 0.36, compared to a distance of 0.30 between the bounding box and the background mask (Right) and a mean distance of 0.97 between the bounding box and generated class prototypes. This suggests the SNN is performing likely matching based on features found in the background rather than on the dorsal fins, reflected in increased model performance whereby using bounding box data to train the SNN sees an increase of 22.94\% top-1, 15.58\% top-5, and 6.53\% top-10 accuracies over using masked data.

By removing all background, the masked SNN is prevented from utilising environmental conditions to aid matching. This finding raises important questions regarding the performance of photo-id aids which do not remove all background before performing matching. If the photo-id catalogue utilised for system evaluation has been collected over a small temporal scale, then results obtained in this experiment suggest that performance may be artificially inflated by the retention of feature heavy background. Further studies will examine the effect of background retention on most likely matching to catalogues gathered over a large temporal scale, as well as the use of out of distribution negative samples \cite{lee_weakly_2022} to train networks robust to the issue of consistent background.


\section{Limitations}\label{sub:Limiations}

One limitation of the system currently is the need to re-train the SNN for each photo-id catalogue. As a result, initial manual curation must be performed before the methodology can be applied. The feasibility of a more general SNN capable of catalogue-agnostic photo-id will be examined in future work. This limitation does not apply to the Mask R-CNN however, which has been found not to require re-training when applied to a new photo-id catalogue regardless of changes in species, geography, or time.

Further, whilst the system has been shown to be robust enough to deal with multiple cetacean species, these have all been dolphins. It is not clear how well the pipeline would perform with cetacean species such as whales or porpoises, or with body parts like flukes. Further studies with catalogues of other species will be explored.


\section{Conclusion}\label{sec:conclusion}

This work examines the use of a pipeline of computer vision models to aid researchers through the curation of photo-id data. The system is capable of operating on raw field images, no pre-processing required, thanks to the use of a dorsal fin detection model. Evaluation of this model shows it is capable of achieving high mAP on photo-id catalogues containing data from different species, collected in different geographical locations, and at different times to the catalogue on which it is trained. Drops in performance are observed when operating over data containing examples of eco-tourism, and further work will examine how best to mitigate this.

Detections are outputted as pixel wise masks, post-processed to improve the chance of catalogue matching. Experimentation to locate the optimal colour threshold confirms erroneous detections can be filtered out with confidence whilst over-exposed fins are retained. 

Outputs are passed to an SNN, trained for the task of most likely catalogue matching, to generate an embedding which is plotted into a latent space. Embeddings are compared to class prototypes using Euclidean distance to generate matches. Evaluation of SNNs trained on processed photo-id data suggests that they are a viable approach to the problem of most likely catalogue matching, and are capable of flagging detections which may be of previously uncatalogued individuals. Experimental results studying the effect of retained background suggest this can negatively impact embedding generation, especially for catalogues collected over a small temporal scale. The use of detection masks negates this effect, preventing embedding generation interference.

{
\balance
\small
\bibliographystyle{IEEEtran}
\bibliography{refs-no-url}
}
\end{document}